\documentclass{article}
\pdfoutput=1

    \usepackage[preprint]{neurips_2019}

\usepackage[utf8]{inputenc} 
\usepackage[T1]{fontenc}    
\usepackage{hyperref}       
\usepackage{url}            
\usepackage{booktabs}       
\usepackage{amsfonts}       
\usepackage{nicefrac}       
\usepackage{microtype}      
\usepackage{bm}
\usepackage{graphicx}
\usepackage[ruled,vlined]{algorithm2e}
\usepackage{authblk}
\usepackage{natbib}[8.31]
\usepackage{float}
\usepackage{caption}
\usepackage{subcaption}
\usepackage[dvipsnames]{xcolor}
\usepackage{wrapfig}
\usepackage{amsmath}
\usepackage{pdfpages}
\usepackage{amssymb}

\def\*#1{\bm{#1}}

\newcommand{\mask}{\ensuremath{\left<\text{mask}\right>}}
\newcommand{\sep}{\ensuremath{\left<\text{sep}\right>}}

\begin{document}

\title{Multi-segment preserving sampling \\ for deep manifold sampler}

\author[1,2]{\textbf{Daniel Berenberg}}
\author[1]{\textbf{Jae Hyeon Lee}}
\author[1]{\textbf{Simon Kelow}}
\author[1]{\textbf{Ji Won Park}}
\author[1]{\textbf{Andrew Watkins}}
\author[1]{\textbf{Vladimir Gligorijevi\'c}}
\author[1]{\textbf{Richard Bonneau}}
\author[1]{\textbf{Stephen Ra}}
\author[1,2,3,4]{\textbf{Kyunghyun Cho}}

\affil[1]{\footnotesize Prescient Design, Genentech}
\affil[2]{\footnotesize Department of Computer Science, Courant Institute of Mathematical Sciences, New York University}
\affil[3]{\footnotesize Center for Data Science, New York University}
\affil[4]{\footnotesize CIFAR Fellow}
\affil[ ]{\texttt{berenberg.daniel@gene.com}}

\maketitle

\vskip -0.15in

\begin{abstract}
Deep generative modeling for biological sequences presents a unique challenge in reconciling the bias-variance trade-off between explicit biological insight and model flexibility.
The deep manifold sampler \citep{gligorijevic2021function} was recently proposed as a means to iteratively sample variable-length protein sequences. 
Sampling was done by exploiting the gradients from a function predictor trained on top of the manifold sampler.
In this work, we introduce an alternative approach to guided sampling that enables the direct inclusion of domain-specific knowledge by designating preserved and non-preserved segments along the input sequence, thereby restricting variation to only select regions.
We call this method ``multi-segment preserving sampling" and present its effectiveness in the context of antibody design.
We train two models: a deep manifold sampler and a GPT-2 language model on nearly six million heavy chain sequences annotated with the \textit{IGHV1-18} gene.
During sampling, we restrict variation to only the complementarity-determining region 3 (CDR3) of the input. We obtain log probability scores from a GPT-2 model for each sampled CDR3 and demonstrate that multi-segment preserving sampling generates reasonable designs while maintaining the desired, preserved regions.
\end{abstract}

\section{Introduction}
Protein sequence families, particularly antibodies, have both well-conserved and variable regions.
In antibodies, the heavy and light chain sequences consist of highly conserved regions known as the framework as well as an array of distinct hypervariable loops, known as complementarity-determining regions (CDRs) \citep{reczko1995prediction}.
Despite the intrinsic variability of CDRs, conditional variation is often conferred by the gene locus admitting the protein \citep{kelows_2020}. Much of an antibody's antigen-binding affinity is owed to the CDRs, while the framework remains fixed or requires minimal change \citep{kuroda2012computer}. For \textit{in silico} modeling, integrating these established aspects of structure and binding can drive the development of better \textit{in situ} antibody therapeutic design \citep{chiu2019antibody}. While work in protein language modeling suggests that models can learn these evolutionary conservation rules~\citep{Rivese2016239118,elnaggar2021prottrans,madani2020progen}, it is an open challenge as to how to explicitly incorporate prior insight at test-time generation, such as sequence-level annotations \citep{anarci2015}, to restrict sampling in certain segments.

The deep manifold sampler was recently proposed as an effective method to sample novel sequences by iterative, optionally gradient-guided steps, of sequence denoising \citep{gligorijevic2021function}.
Empirically, gradient-based guided sampling was shown to selectively encourage changes in functional sites, implicitly leaving non-functional regions unperturbed.
In this work, we propose an alternative to the gradient-based guided design procedure in which predefined regions of a sequence are explicitly preserved, leaving sampling to take place in \textit{a priori} known notable sequence regions.
We conduct an experiment on antibody sequences to demonstrate the deep manifold sampler's ability to focus sampling on a subset of sequence positions. We do so by deliberately corrupting select regions of antibody sequences, that correspond to CDRs, and evaluating the length distribution and composition of sampled CDRs.

\section{Background: the Deep Manifold Sampler} \label{sec:background}

The deep manifold sampler \citep{gligorijevic2021function} is a denoising autoencoder (DAE) specialized for handling variable-length sequences.
As with a typical DAE \citep{vincent2008extracting}, the deep manifold sampler consists of three modules; a corruption process $C(\tilde{x}|x)$, an encoder $F$ and a decoder $G$. Unlike the usual DAE however, the deep manifold sampler has an extra module that determines the change in the length, which we call the ``length conversion'' \citep{shu2020latent}. 

The deep manifold sampler assumes as input a sequence of discrete tokens, $x=(x_1, x_2, \ldots, x_{|x|})$, where each token $x_t$ is an item from a finite vocabulary $V$ of unique words or subwords. In the case of protein sequence modeling, $V$ consists of all unique amino acids. The sequence $x$ is corrupted with the corruption process $C$, resulting in a {\it noisy} input sequence $\tilde{x} \sim C(\tilde{x}|x)$. This corruption process can be arbitrary as long as it is largely local and unstructured. It may even alter the length of the sequence, $|x| \neq |\tilde{x}|$. 

The encoder $F$ turns the corrupted sequence $\tilde{x}$ into a set of hidden vectors, $h=(h_1, h_2, \ldots, h_{|\tilde{x}|})$, where $h_t \in \mathbb{R}^d$. The encoder can be implemented using any of the widely-used deep architectures, such as transformers \citep{vaswani2017attention}, convolutional networks \citep{gehring2017convolutional} and recurrent networks \citep{sutskever2014sequence,bahdanau2014neural}. In this work, we follow the original deep manifold sampler's encoder, which was implemented as a transformer.

The hidden vectors are pooled to form a single-vector representation:
\begin{align*}
    \bar{h} = \frac{1}{|\tilde{x}|} \sum_{t=1}^{|\tilde{x}|} h_t.
\end{align*} 
This pooled representation is used by the length conversion to predict the change in the length. At training time, this length change predictor is trained to output $\Delta l^* = |\tilde{x}| - |x|$. When we sample sequences from the deep manifold sampler after training, we use the predicted change $\Delta l$ to adjust the size of the hidden vector set. The adjusted hidden vector set consists of $|\tilde{x}|+\Delta l$ hidden vectors, $z=(z_1, \ldots, z_{|\tilde{x}|+\Delta l})$, where each vector is a weighted sum of the previous hidden vectors $h_1, h_2, \cdots, h_{|\tilde x|}$. To wit, we define
\begin{align}
\label{eq:length-conversion}
z_t = \sum_{t'=1}^{|\tilde{x}|} \omega_{t,t'} h_{t'}
\end{align}
with the position-based softmax weights $w_{t, t'}$ preferring $h_{t'}$ closest to the length-scaled position $|\tilde x|/(|\tilde x| + \Delta l)t$, as follows:
\begin{align}
w_{t, t'} &= \frac{\exp(q_{t, t'})}{\sum_{t''=1}^{|\tilde x|} \exp(q_{t, t''})}\\ \label{eq:sigma}
q_{t, t'} &\propto 
\frac{-1}{2\sigma^2}
\left(  t' - \frac{|\tilde{x}|}{|\tilde{x}| + \Delta l}t \right)^2.
\end{align}
In Eq.~\eqref{eq:sigma}, $\sigma$ is a learned smoothing parameter.
The decoder $G$ then takes this transformed hidden vector sequence $z$ and outputs a corresponding sequence of logit vectors, $\tilde{y}=(\tilde{y}_1, \ldots, \tilde{y}_{|\tilde{x}|+\Delta l})$, where $\tilde{y}_t \in \mathbb{R}^{|V|}$. These logits are turned into probability distributions over the vocabulary $V$ in many different ways. The original deep manifold sampler implements a non-autoregressive approach \citep{gu2017non,lee2018deterministic}, where each logit is independently turned into a distribution:

\begin{equation}
    \label{eq:non-ar}
    p(y_t = v | \tilde{x}, \Delta l) = \frac{\exp \left( \tilde{y}_t^v + b^v \right)}{\sum_{v' \in V} \exp \left( \tilde{y}_t^{v'} + b^{v'} \right)},
\end{equation}
where $b^v$ is a bias for token $v$.

It is, however, also possible to use these logits together with a more powerful output module, such as a conditional random field (CRF; \citealt{lafferty2001}), as was recently done in \citep{yi2021nettime}, and autoregressive language models~\citep{mikolov2010recurrent}. For experiments in this paper, we use a variant of the deep manifold sampler with a CRF at the end of the decoder. 

At training time, we minimize the negative log-probability of the original sequence $x$ given the corrupted version $\tilde{x}$ and the known $\Delta l^*$ to train the encoder and decoder, while minimizing the negative log-probability of $\Delta l^*$ to train the length change predictor. We parameterize the latter as a classifier. Once training is done, we can draw a series of samples from the deep manifold sampler by repeating the process of corruption, length conversion, and reconstruction. 

While the original deep manifold sampler has an additional function predictor that can be used to guide the sampling procedure, we omit that here, as this is optional and can be replaced with another computational oracle without altering the sampling procedure that is the focus of this paper.

\section{Multi-Segment Preserving Sampling}

The deep manifold sampler was originally proposed in the context of protein design in \citet{gligorijevic2021function}. Within this setting, we often consider biological, chemical, and physical knowledge in order to impose constraints that narrow down a large, combinatorial search space \citep{street1999computational,woolfson2021brief}. The deep manifold sampler, on the other hand, stays true to the key principle of deep learning, that is, end-to-end learning, which makes it challenging to explicitly incorporate this knowledge into both learning and sampling. In this paper, we take one step towards enabling this into the sampling procedure of the deep manifold sampler. We assume the availability of knowledge in which segments of an original sequence, from which sampling starts, must be preserved in order to maintain a set of desirable properties. For example, in the case of antibody engineering, it may be desirable to only alter CDR loops while leaving all framework residues intact \citep{kuroda2012computer}.

\begin{figure}[!t]
  \centering
  \includegraphics[width=\textwidth]{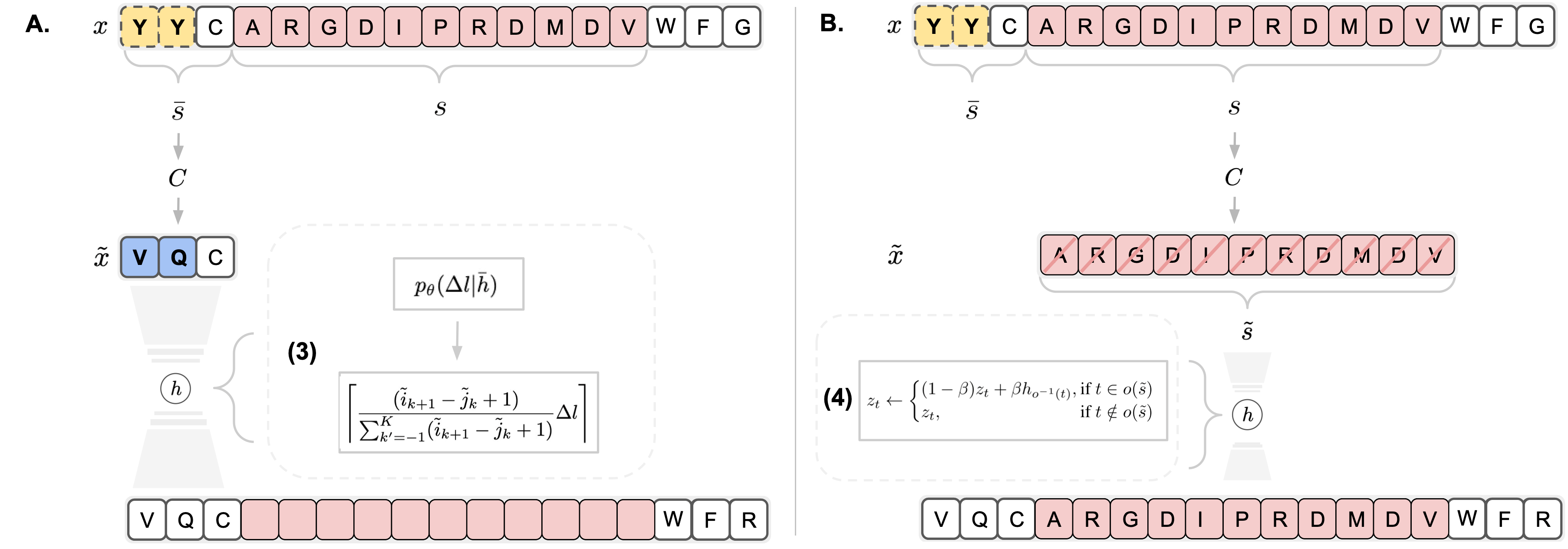}
  \caption{Multi-segment preserving sampling. (A) Non-preserved segments $\bar{s}$ are corrupted using corruption process $C$, for which a given token (yellow) may be randomly perturbed (blue). This is encoded into the hidden vector set $h$. Length change predictor $p_{\theta}(\Delta l|\bar{h})$ outputs $\Delta l$, which is distributed across $\bar{s}$ (Eq. \ref{eq:nonpreserved}).  (B) Segment-preserving sampling follows similar operations on preserved segment $s$ (red) with notable differences. Corruption $C$ yields an  unaltered sequence $\tilde{x}$ and we carry over hidden vector $h_t$ of a token within preserved segment $\tilde{s}$ with strength $\beta$ (Eq. \ref{eq:preserved}).} 
  \label{fig:overview}
\end{figure}

Let $x = (x_1, x_2, \ldots, x_{|x|})$ be the initial sequence from which we run the deep manifold sampler to draw a series of samples over the sequence manifold. Instead of unconstrained sampling, we consider a scenario in which we are provided with a set of non-overlapping segments of the sequence that must be preserved in-order by their starting and ending indices (inclusive):

\begin{align*}
    s = ((i_1, j_1), \ldots, (i_K, j_K))
\end{align*}

subject to $i_1>0$, $i_k \leq j_k$ for all $k$, $j_k < i_{k+1}$ for all $k$, and $j_K < |x|-1$. We refer to this set as a {\it preserved-segment set}. Likewise, we can imagine the complement segment set $\bar{s}$ that contains all the segments that are between the to-be-preserved segments in $s$:

\begin{align*} 
    \bar{s} = ((0, i_1-1), (j_1+1, i_2-1), \ldots, (j_K+1, |x|-1)).
\end{align*}

In order to preserve these segments while altering the remaining parts of the sequence, including their respective lengths, we make a series of modifications to the sampling procedure of the deep manifold sampler. First, we alter the corruption process $C$ such that it does not corrupt the preserved segments. For instance, if the corruption process randomly adds or removes tokens, this is only done to the segments in the complement set $\bar{s}$ but not to those in $s$. The corrupted sequence $\tilde x$ contains an indexing change due to insertions and deletions, so the description of the segment set $s$ must be updated to reflect this --- we denote the preserved segment set of $\tilde x$ by $\tilde{s}$. 

The encoder still encodes $\tilde{x}$ into the hidden vector set $h$, as described in Section \ref{sec:background}. While the length change prediction steps also stay the same, the returned length change $\Delta l$ needs to be distributed across the non-preserved segments in order to avoid altering the length of any preserved segment in $\tilde{s}$. We do so proportional to the original lengths of the non-preserved segments. Concretely, we add to the length of each non-preserved segment $(\tilde{j}_k+1, \tilde{i}_{k+1}-1)$:

\begin{equation}
    \label{eq:nonpreserved}
    \left\lceil \frac{(\tilde{i}_{k+1} - \tilde{j}_k+1)} 
    {\sum_{k=-1}^{K} (\tilde{i}_{k+1} - \tilde{j}_k+1)}
    \Delta l \right\rceil,
\end{equation}
where $\tilde{j}_0 = 0$ and $\tilde{i}_{K+1} = |\tilde{x}|-1$. 

After distributing the length difference among the non-preserved segments, we can now construct the index map $o$ that tells us which segment in the new sequence corresponds to each of the preserved segment in $\tilde{x}$. In other words, $y_{o(\tilde{i}_k):o(\tilde{j}_k)} = \tilde{x}_{\tilde{i}_k:\tilde{j}_k}$. Let us use $o(\tilde{s})$ to denote the preserved-segment set derived from $s$ and the length distribution above. 

The actual length conversion happens just like before, as in Eq.~\eqref{eq:length-conversion}. We however add an extra step after the length conversion in order to give the decoder a hint about preserved segments and their contents. This is done by carrying over the original hidden vector $h_t$ of a token within a preserved segment:

\begin{equation}
    \label{eq:preserved}
    z_t \leftarrow
    \begin{cases}
        (1-\beta) z_t + \beta h_{o^{-1}(t)}, &\text{if } t \in o(\tilde{s}) \\
        z_t, &\text{if } t \notin o(\tilde{s}) \\
    \end{cases}
\end{equation}

$o^{-1}$ is the inverse index map, and $\beta \in [0, 1]$ is the strength of carry-over. 

The decoder turns this length-converted and segment-preserving hidden sequence $z$ into a sequence of logit vectors $\tilde{y}$, just like the original sampling procedure. We then modify the logit vector corresponding to a token with a preserved segment to force the sampled outcome to preserve the token identity:

\begin{equation}
    \label{eq:logits}
    \tilde{y}_t^v
    \leftarrow
    \begin{cases}
        \infty, &\text{ if } t \in o(\tilde{s}) \text{ and } v = \tilde{x}_{o^{-1}(t)} \\
        -\infty, &\text{ if } t \in o(\tilde{s}) \text{ and } v \neq \tilde{x}_{o^{-1}(t)} \\
        \tilde{y}_t^v, &\text{ if } t \notin o(\tilde{s})
    \end{cases}
\end{equation}

In the case of non-autoregressive modeling, this would result in a Categorical distribution for a preserved token to assign the entire probability mass ($=1$) to the original token identity. If a CRF is used at the end, this would prevent any sequence that violates preservation from being decoded out with non-zero probability. 

We can repeat this sampling step with the newly sampled sequence and the corresponding preserved-segment set. This allows us to iteratively draw a series of samples while preserving the segments from the original sequence, designated by the preserved-segment set $s$. Because this iterative sampling procedure preserves multiple segments and their contents, we refer to this procedure as {\it multi-segment preserving sampling} (Figure \ref{fig:overview}).

\section{Related Work}

There are two alternative sequence-modeling paradigms that are closely related to the deep manifold sampler. We briefly describe each of them here in the context of whether and how multi-segment preservation can be implemented.  

\subsection{Masked Language Models}

A masked language model is a special case of a DAE, similar to the deep manifold sampler \citep{devlin2018bert,liu2019roberta}. A corruption process $C$ in a masked language model is designed to apply one of three types of corruption: (1) replace a token with a special {\mask} token, (2) replace a token with another, randomly-selected token, and (3) no alteration, to a random subset of the tokens within each sequence. All of these types do {\it not} alter the length of the sequence. The masked language model then reconstructs only the corrupted subset, rather than the full sequence as in the deep manifold sampler. 

Masked language modeling was originally motivated as a way to pretrain a large-scale neural network rather than as a generative model from which to draw sequences. While \citet{wang2019bert} and \citet{goyal2021exposing} demonstrated that masked language models can yield well-formed sequences, they are neither as popular nor applicable as as other models --- such as autoregressive language models or the deep manifold sampler --- for sequence generation because they do not have a mechanism to automatically model the length distribution.

With the caveat that the length of a sequence cannot be altered at sampling time, a masked language model can be used for multi-segment preserving sampling. This can still be useful as a ``plug-and-play" proposal distribution for a number of downstream tasks including rational design and directed evolution of proteins \citep{woolfson2021brief,arnold1998design,yang2019machine,meier2021language}. However, this approach is limited compared to the proposed strategy of multi-segment preserving sampling with the deep manifold sampler, as our proposal is able to both dynamically adapt the length of a sequence and preserve segments.  

\subsection{Denoising Sequence-to-Sequence Models}

A denoising sequence-to-sequence (Seq2Seq) model, where the decoder is autoregressive, has been studied previously in the context of natural language processing \citep{hill2016learning,lewis2019bart}. Unlike masked language models, and similar to the deep manifold sampler, the denoising Seq2Seq model can adaptively change the length of a sequence. However, unlike the deep manifold sampler, there is less control over the direct manipulation of the intermediate hidden vectors, as some of the dependencies between the tokens in the output sequence are captured directly by the decoder without relying on the intermediate representation between the encoder and decoder. It is however an interesting future direction to compare the denoising Seq2Seq model against the deep manifold sampler.

With respect to multi-segment preservation, denoising Seq2Seq models do not readily admit such a sampling strategy. This is due to the inherent intractability in decoding from an autoregressive model with unbounded context. This intractability is often addressed by approximate decoding, such as beam search, which is known to have suboptimal behaviors despite its successful and wide use \citep{welleck2020consistency,welleck2019neural}. 
Several studies have proposed to extend beam search to incorporate such constraints \citep{hokamp2017lexically,post2018fast}. Unfortunately most of these approaches incur great computational cost, as their computational complexity grows often linearly with respect to the number of preserved segments --- or the beam size must grow accordingly --- because the underlying algorithm decodes in a greedy left-to-right fashion with a limited hypothesis set (beam). While it is possible to modify a denoising Seq2Seq model to admit multi-segment preserving sampling by letting the decoder only reconstruct non-preserved segments, this is out of scope for this paper.

\section{Experiments}

The proposed algorithm for multi-segment preserving sampling is designed to completely preserve designated segments. Here, we demonstrate a potential application in antibody design enabled by our algorithm coupled with the deep manifold sampler. Antibodies with a particular V-gene have fixed lengths in the framework as well as in the CDR1 and CDR2 regions. As a result, antibodies display most of their diversity in length and amino acid composition in CDR3 \citep{glanville2009precise}. To demonstrate the effectiveness of our approach and restricted variation of the preserved segments, we select all unique human antibody sequences with the \textit{IGHV1-18} gene from the Observed Antibody Space (OAS) database \citep{olsen2022observed} for multi-segment preserving sampling. Using a deep manifold sampler, we sample exclusively from the CDR3, while preserving other regions, and show the length and log-probability (GPT-2) distributions of the generated sequences qualitatively coincide with that of the test data. Table \ref{tab:tab1} illustrates examples of sampled CDR3 regions under different settings of carry-over strength $\beta$.

\begin{table}
\centering
\begin{tabular}{c|l|c}
     \textbf{$\beta$} & \textbf{Aligned CDR3 sequence} & \textbf{Edit distance}\\
     \midrule
     N/A (original) & \texttt{ARDPEWDPF-QANY-YYYGMDV}
 & 0 \\
     0.0 & \texttt{ARDPEWDPF-QAN--YYYGMDV} & 3 \\
     0.1 & \texttt{ARDPEWDPFFQANYNYYYGMVD} & 3 \\
     0.5 & \texttt{KRDPEWDRF-QAPY-YTVGMDV} & 5 \\
     0.9 & \texttt{ARGPECDPH-QAV-DIYYGMDV} & 6 \\

\vspace{.2cm}
\end{tabular}
\caption{Example outputs of multi-segment preserving sampling when restricting variation to the CDR3 region under different settings of $\beta$. Display is restricted to the sampled region, the rest is preserved by construction.}
\label{tab:tab1}
\end{table}

\subsection{Training details}
We obtained 5,971,552 unique human antibody heavy chain sequences with the \textit{IGHV1-18} gene from the OAS database, with 2,000 and 10,000 sequences set aside for validation and test sets respectively and the remaining used for training.\footnote{
Only sequences with "Redundancy $>$ 1" were retained.
} 
We trained a deep manifold sampler on the training set with a constant learning rate of $10^{-4}$ for 60K mini-batch steps with the batch size of 128. The model consisted of a two-layer transformer encoder and decoder, each with 8 heads and the total embedding dimension of 256 and feed-forward layer dimension of 1024. The last layer consists of a CRF for final sequence generation. The rest of the training procedure was the same as described in \citet{gligorijevic2021function}.

In addition, we also trained an autoregressive GPT-2 model using HuggingFace Transformers library v4.16.2 \citep{wolf-etal-2020-transformers} on the same training set in order to demonstrate that the sampler-generated sequences capture the amino-acid token distribution observed in the training set. The model consisted of 6 attention layers with 8 heads and a total embedding dimension of 512 and was trained with a constant learning rate of $4 \times 10^{-4}$ for 25K mini-batch steps with batch size of 1024. The other parameters were set to the default values provided by the package.

\subsection{Sampling details and results}
For each sequence in the test set, we applied multi-segment preserving sampling for one iteration, preserving all non-CDR3 regions with four different $\beta$ values of 0, 0.1, 0.5, and 0.9.\footnote{
The region annotation was obtained from the OAS data unit files.
}

\begin{figure}[!t]
  \centering
  \includegraphics[width=\textwidth]{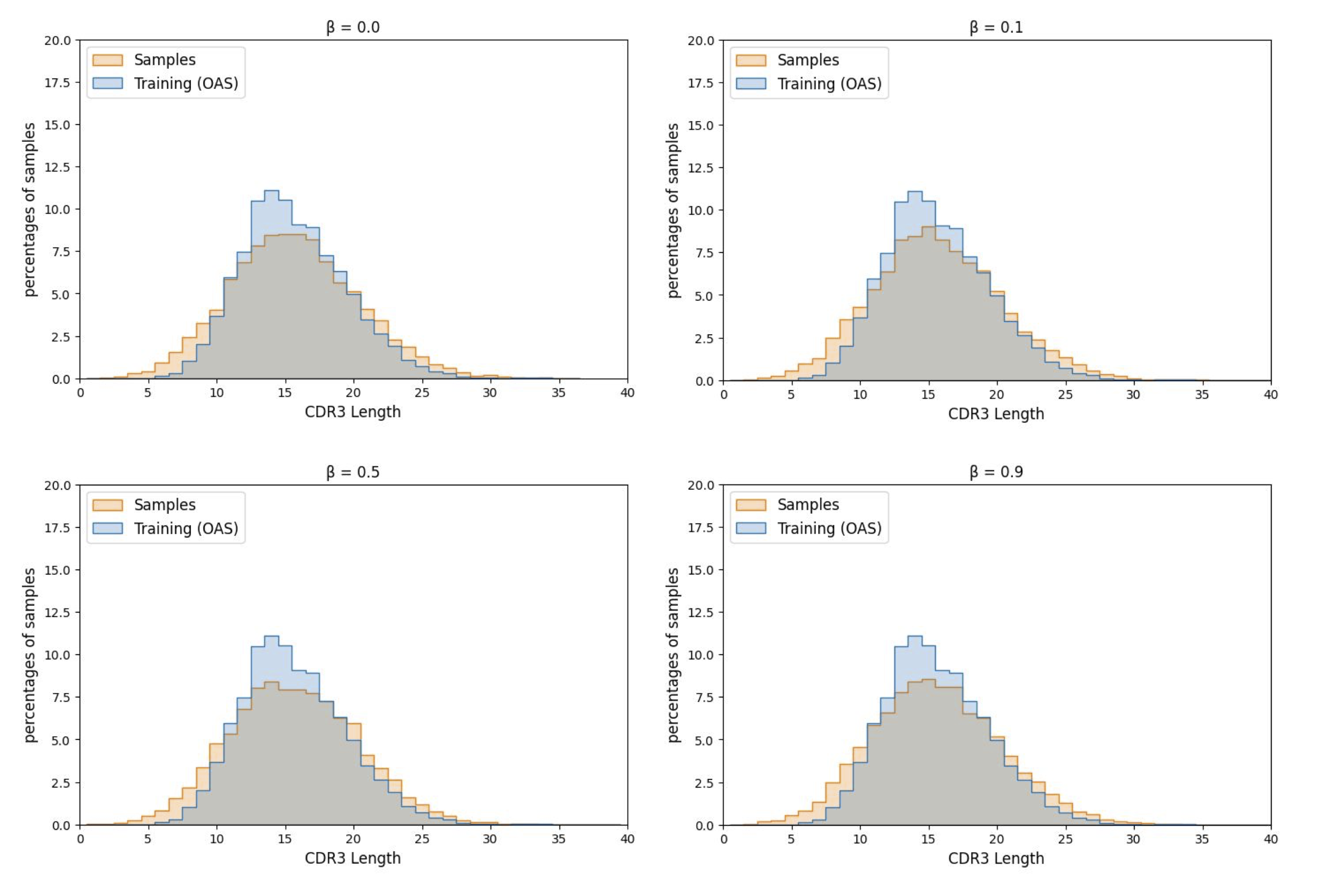}
  \caption{The normalized distribution of the CDR3 lengths of the deep manifold sampler-generated sequences (``Samples") and the test set sequences (``Training (OAS)") with four different $\beta$ parameters. From Top Left, Clockwise: Samples were generated with $\beta = 0, 0.1, 0.9$, and $0.5$.} 
  \label{fig:cd3}
\end{figure}

\begin{figure}[!t]
  \centering
  \includegraphics[width=\textwidth]{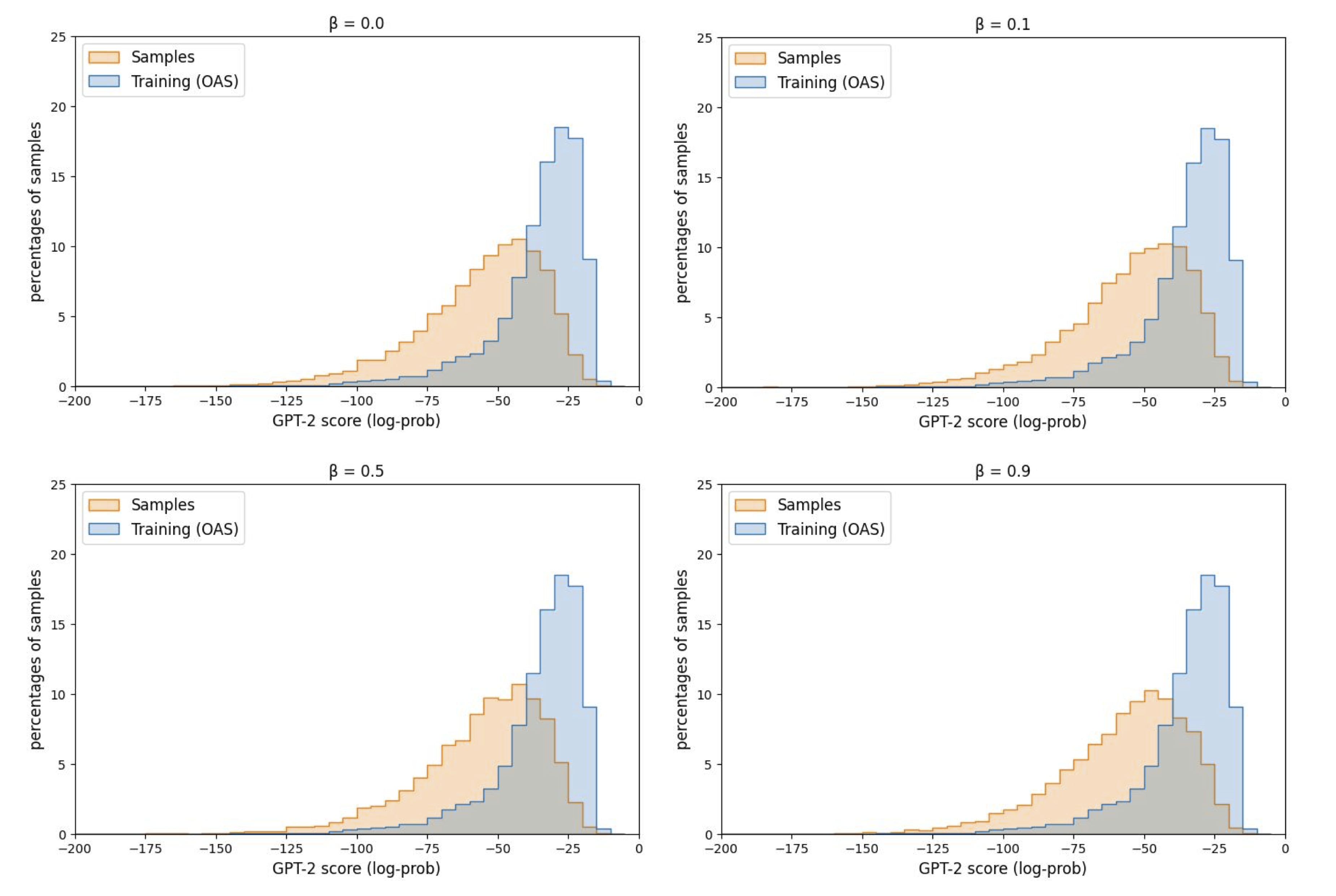}
  \caption{The normalized distribution of the GPT-2 scores of the deep manifold sampler-generated sequences (``Samples") and the test set sequences (``Training (OAS)") with four different $\beta$ parameters. From Top Left, Clockwise: Samples were generated with $\beta = 0, 0.1, 0.9$, and $0.5$.} 
  \label{fig:gpt}
\end{figure}

Figure \ref{fig:cd3} and Figure \ref{fig:gpt} show the length and log-probability (GPT-2) distributions of the generated sequences with changes in CDR3 and the test data across all selected $\beta$ values. The CDR3 length distribution of the generated samples matches the natural sequence length distribution for each value of $\beta$. The GPT-2 log-probability distribution of the samples has lower overall mean compared to that of the test distribution but is still within the same range, indicating that the samples are plausible. Both distributions vary only slightly with different values of $\beta$. These two results show the effectiveness of the sampling strategy for generating diverse antibody sequences, restricted to user-defined regions.

\begin{figure}[!t]
  \centering
  \includegraphics[scale=0.5]{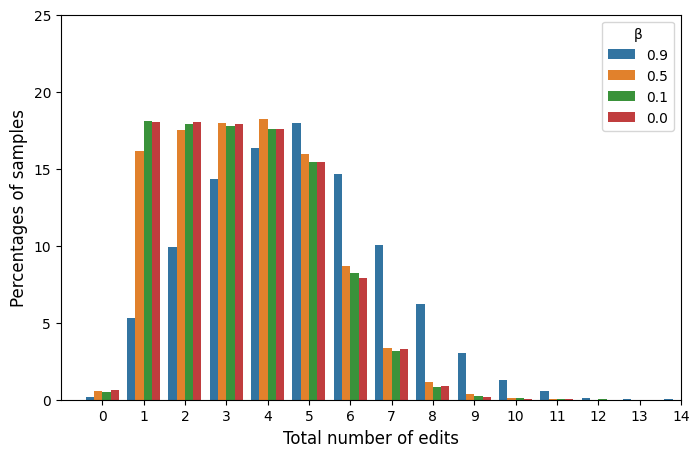}
  \caption{The distribution of edit distances between generated samples and their seed sequences with varying settings of $\beta$ (0, 0.1, 0.5, and 0.9).}
  \label{fig:edits}
\end{figure}

In Figure \ref{fig:edits}, we illustrate the distribution of the number of edits in the generated sequences relative to the input seed sequences, including substitutions, insertions, and deletions. The distributional mean increases slightly with higher values of $\beta$. For future work, we plan on a more systematic understanding of the effects of carry-over strength $\beta$ on sample quality and diversity.



\section{Conclusion}
We have proposed a sampling procedure for the deep manifold sampler that explicitly preserves designated segments of the input sequence, allowing variation to occur only in non-preserved regions. We find that this approach, multi-segment preserving sampling, is applicable to a number of design problems in the life sciences where we often want to use prior knowledge made available in these domains. With biological sequence design for example, we want to sample new, diverse designs that avoid perturbing well-conserved regions of the input.  In this way, we demonstrate the utility of multi-segment preserving sampling by restricting sampling to the CDR3 regions of a collection of antibody heavy chains with the \textit{IGHV1-18} gene and validating the resulting CDR3 designs against a separate GPT-2 model. As shown in Figure \ref{fig:gpt}, the sampled CDR3 regions admit high log-probability scores by the GPT-2 model, providing evidence that the samples are plausible.
Additionally, the CDR3 length distribution of the samples traces the observed length distribution in the training set, suggesting that the model adequately captures the variability in non-preserved segment lengths, despite the lack of explicit provision during training.
In future work, we will extend our exploration on the effect of the carry-over strength $\beta$ in terms of sample quality as well as its usage in conjunction with the function predictor for guided sampling proposed in \citet{gligorijevic2021function}.

\newpage


\vskip 0.2in
\bibliographystyle{plainnat}
\bibliography{references}

\begin{thebibliography}{38}
\providecommand{\natexlab}[1]{#1}
\providecommand{\url}[1]{\texttt{#1}}
\expandafter\ifx\csname urlstyle\endcsname\relax
  \providecommand{\doi}[1]{doi: #1}\else
  \providecommand{\doi}{doi: \begingroup \urlstyle{rm}\Url}\fi

\bibitem[Arnold(1998)]{arnold1998design}
Frances~H Arnold.
\newblock Design by directed evolution.
\newblock \emph{Accounts of chemical research}, 31\penalty0 (3):\penalty0
  125--131, 1998.

\bibitem[Bahdanau et~al.(2014)Bahdanau, Cho, and Bengio]{bahdanau2014neural}
Dzmitry Bahdanau, Kyunghyun Cho, and Yoshua Bengio.
\newblock Neural machine translation by jointly learning to align and
  translate.
\newblock \emph{arXiv preprint arXiv:1409.0473}, 2014.

\bibitem[Chiu et~al.(2019)Chiu, Goulet, Teplyakov, and
  Gilliland]{chiu2019antibody}
Mark~L Chiu, Dennis~R Goulet, Alexey Teplyakov, and Gary~L Gilliland.
\newblock Antibody structure and function: the basis for engineering
  therapeutics.
\newblock \emph{Antibodies}, 8\penalty0 (4):\penalty0 55, 2019.

\bibitem[Devlin et~al.(2018)Devlin, Chang, Lee, and Toutanova]{devlin2018bert}
Jacob Devlin, Ming-Wei Chang, Kenton Lee, and Kristina Toutanova.
\newblock Bert: Pre-training of deep bidirectional transformers for language
  understanding.
\newblock \emph{arXiv preprint arXiv:1810.04805}, 2018.

\bibitem[Dunbar and Deane(2015)]{anarci2015}
James Dunbar and Charlotte~M. Deane.
\newblock {ANARCI: antigen receptor numbering and receptor classification}.
\newblock \emph{Bioinformatics}, 32\penalty0 (2):\penalty0 298--300, 09 2015.
\newblock ISSN 1367-4803.
\newblock \doi{10.1093/bioinformatics/btv552}.
\newblock URL \url{https://doi.org/10.1093/bioinformatics/btv552}.

\bibitem[Elnaggar et~al.(2021)Elnaggar, Heinzinger, Dallago, Rihawi, Wang,
  Jones, Gibbs, Feher, Angerer, Steinegger, Bhowmik, and
  Rost]{elnaggar2021prottrans}
Ahmed Elnaggar, Michael Heinzinger, Christian Dallago, Ghalia Rihawi, Yu~Wang,
  Llion Jones, Tom Gibbs, Tamas Feher, Christoph Angerer, Martin Steinegger,
  Debsindhu Bhowmik, and Burkhard Rost.
\newblock Prottrans: Towards cracking the language of life's code through
  self-supervised deep learning and high performance computing, 2021.

\bibitem[Gehring et~al.(2017)Gehring, Auli, Grangier, Yarats, and
  Dauphin]{gehring2017convolutional}
Jonas Gehring, Michael Auli, David Grangier, Denis Yarats, and Yann~N Dauphin.
\newblock Convolutional sequence to sequence learning.
\newblock In \emph{International Conference on Machine Learning}, pages
  1243--1252. PMLR, 2017.

\bibitem[Glanville et~al.(2009)Glanville, Zhai, Berka, Telman, Huerta, Mehta,
  Ni, Mei, Sundar, Day, et~al.]{glanville2009precise}
Jacob Glanville, Wenwu Zhai, Jan Berka, Dilduz Telman, Gabriella Huerta,
  Gautam~R Mehta, Irene Ni, Li~Mei, Purnima~D Sundar, Giles~MR Day, et~al.
\newblock Precise determination of the diversity of a combinatorial antibody
  library gives insight into the human immunoglobulin repertoire.
\newblock \emph{Proceedings of the National Academy of Sciences}, 106\penalty0
  (48):\penalty0 20216--20221, 2009.

\bibitem[Gligorijević et~al.(2021)Gligorijević, Berenberg, Ra, Watkins,
  Kelow, Cho, and Bonneau]{gligorijevic2021function}
Vladimir Gligorijević, Daniel Berenberg, Stephen Ra, Andrew Watkins, Simon
  Kelow, Kyunghyun Cho, and Richard Bonneau.
\newblock Function-guided protein design by deep manifold sampling.
\newblock \emph{bioRxiv}, 2021.

\bibitem[Goyal et~al.(2021)Goyal, Dyer, and
  Berg-Kirkpatrick]{goyal2021exposing}
Kartik Goyal, Chris Dyer, and Taylor Berg-Kirkpatrick.
\newblock Exposing the implicit energy networks behind masked language models
  via metropolis--hastings.
\newblock \emph{arXiv preprint arXiv:2106.02736}, 2021.

\bibitem[Gu et~al.(2017)Gu, Bradbury, Xiong, Li, and Socher]{gu2017non}
Jiatao Gu, James Bradbury, Caiming Xiong, Victor~OK Li, and Richard Socher.
\newblock Non-autoregressive neural machine translation.
\newblock \emph{arXiv preprint arXiv:1711.02281}, 2017.

\bibitem[Hill et~al.(2016)Hill, Cho, and Korhonen]{hill2016learning}
Felix Hill, Kyunghyun Cho, and Anna Korhonen.
\newblock Learning distributed representations of sentences from unlabelled
  data.
\newblock \emph{arXiv preprint arXiv:1602.03483}, 2016.

\bibitem[Hokamp and Liu(2017)]{hokamp2017lexically}
Chris Hokamp and Qun Liu.
\newblock Lexically constrained decoding for sequence generation using grid
  beam search.
\newblock \emph{arXiv preprint arXiv:1704.07138}, 2017.

\bibitem[Kelow et~al.(2020)Kelow, Adolf-Bryfogle, and Dunbrack]{kelows_2020}
Simon~P. Kelow, Jared Adolf-Bryfogle, and Roland~L. Dunbrack.
\newblock Hiding in plain sight: structure and sequence analysis reveals the
  importance of the antibody de loop for antibody-antigen binding.
\newblock \emph{mAbs}, 12\penalty0 (1):\penalty0 1840005, 2020.
\newblock \doi{10.1080/19420862.2020.1840005}.
\newblock URL \url{https://doi.org/10.1080/19420862.2020.1840005}.
\newblock PMID: 33180672.

\bibitem[Kuroda et~al.(2012)Kuroda, Shirai, Jacobson, and
  Nakamura]{kuroda2012computer}
Daisuke Kuroda, Hiroki Shirai, Matthew~P Jacobson, and Haruki Nakamura.
\newblock Computer-aided antibody design.
\newblock \emph{Protein engineering, design \& selection}, 25\penalty0
  (10):\penalty0 507--522, 2012.

\bibitem[Lafferty et~al.(2001)Lafferty, McCallum, and Pereira]{lafferty2001}
John~D. Lafferty, Andrew McCallum, and Fernando C.~N. Pereira.
\newblock Conditional random fields: Probabilistic models for segmenting and
  labeling sequence data.
\newblock In \emph{Proceedings of the Eighteenth International Conference on
  Machine Learning}, ICML '01, page 282–289, San Francisco, CA, USA, 2001.
  Morgan Kaufmann Publishers Inc.
\newblock ISBN 1558607781.

\bibitem[Lee et~al.(2018)Lee, Mansimov, and Cho]{lee2018deterministic}
Jason Lee, Elman Mansimov, and Kyunghyun Cho.
\newblock Deterministic non-autoregressive neural sequence modeling by
  iterative refinement.
\newblock \emph{arXiv preprint arXiv:1802.06901}, 2018.

\bibitem[Lewis et~al.(2019)Lewis, Liu, Goyal, Ghazvininejad, Mohamed, Levy,
  Stoyanov, and Zettlemoyer]{lewis2019bart}
Mike Lewis, Yinhan Liu, Naman Goyal, Marjan Ghazvininejad, Abdelrahman Mohamed,
  Omer Levy, Ves Stoyanov, and Luke Zettlemoyer.
\newblock Bart: Denoising sequence-to-sequence pre-training for natural
  language generation, translation, and comprehension.
\newblock \emph{arXiv preprint arXiv:1910.13461}, 2019.

\bibitem[Liu et~al.(2019)Liu, Ott, Goyal, Du, Joshi, Chen, Levy, Lewis,
  Zettlemoyer, and Stoyanov]{liu2019roberta}
Yinhan Liu, Myle Ott, Naman Goyal, Jingfei Du, Mandar Joshi, Danqi Chen, Omer
  Levy, Mike Lewis, Luke Zettlemoyer, and Veselin Stoyanov.
\newblock Roberta: A robustly optimized bert pretraining approach.
\newblock \emph{arXiv preprint arXiv:1907.11692}, 2019.

\bibitem[Madani et~al.(2020)Madani, McCann, Naik, Keskar, Anand, Eguchi, Huang,
  and Socher]{madani2020progen}
Ali Madani, Bryan McCann, Nikhil Naik, Nitish~Shirish Keskar, Namrata Anand,
  Raphael~R. Eguchi, Po-Ssu Huang, and Richard Socher.
\newblock Progen: Language modeling for protein generation, 2020.

\bibitem[Meier et~al.(2021)Meier, Rao, Verkuil, Liu, Sercu, and
  Rives]{meier2021language}
Joshua Meier, Roshan Rao, Robert Verkuil, Jason Liu, Tom Sercu, and Alex Rives.
\newblock Language models enable zero-shot prediction of the effects of
  mutations on protein function.
\newblock \emph{Advances in Neural Information Processing Systems}, 34, 2021.

\bibitem[Mikolov et~al.(2010)Mikolov, Karafi{\'a}t, Burget, Cernock{\`y}, and
  Khudanpur]{mikolov2010recurrent}
Tomas Mikolov, Martin Karafi{\'a}t, Lukas Burget, Jan Cernock{\`y}, and Sanjeev
  Khudanpur.
\newblock Recurrent neural network based language model.
\newblock In \emph{Interspeech}, volume~2, pages 1045--1048. Makuhari, 2010.

\bibitem[Olsen et~al.(2022)Olsen, Boyles, and Deane]{olsen2022observed}
Tobias~H Olsen, Fergus Boyles, and Charlotte~M Deane.
\newblock Observed antibody space: A diverse database of cleaned, annotated,
  and translated unpaired and paired antibody sequences.
\newblock \emph{Protein Science}, 31\penalty0 (1):\penalty0 141--146, 2022.

\bibitem[Post and Vilar(2018)]{post2018fast}
Matt Post and David Vilar.
\newblock Fast lexically constrained decoding with dynamic beam allocation for
  neural machine translation.
\newblock \emph{arXiv preprint arXiv:1804.06609}, 2018.

\bibitem[Reczko et~al.(1995)Reczko, Martin, Bohr, and
  Suhai]{reczko1995prediction}
Martin Reczko, Andrew~CR Martin, Henrik Bohr, and S{\'a}ndor Suhai.
\newblock Prediction of hypervariable cdr-h3 loop structures in antibodies.
\newblock \emph{Protein Engineering, Design and Selection}, 8\penalty0
  (4):\penalty0 389--395, 1995.

\bibitem[Rives et~al.(2021)Rives, Meier, Sercu, Goyal, Lin, Liu, Guo, Ott,
  Zitnick, Ma, and Fergus]{Rivese2016239118}
Alexander Rives, Joshua Meier, Tom Sercu, Siddharth Goyal, Zeming Lin, Jason
  Liu, Demi Guo, Myle Ott, C.~Lawrence Zitnick, Jerry Ma, and Rob Fergus.
\newblock Biological structure and function emerge from scaling unsupervised
  learning to 250 million protein sequences.
\newblock \emph{Proceedings of the National Academy of Sciences}, 118\penalty0
  (15), 2021.
\newblock ISSN 0027-8424.
\newblock \doi{10.1073/pnas.2016239118}.
\newblock URL \url{https://www.pnas.org/content/118/15/e2016239118}.

\bibitem[Shu et~al.(2020)Shu, Lee, Nakayama, and Cho]{shu2020latent}
Raphael Shu, Jason Lee, Hideki Nakayama, and Kyunghyun Cho.
\newblock Latent-variable non-autoregressive neural machine translation with
  deterministic inference using a delta posterior.
\newblock In \emph{Proceedings of the AAAI Conference on Artificial
  Intelligence}, volume~34, pages 8846--8853, 2020.

\bibitem[Street and Mayo(1999)]{street1999computational}
Arthur~G Street and Stephen~L Mayo.
\newblock Computational protein design.
\newblock \emph{Structure}, 7\penalty0 (5):\penalty0 R105--R109, 1999.

\bibitem[Sutskever et~al.(2014)Sutskever, Vinyals, and
  Le]{sutskever2014sequence}
Ilya Sutskever, Oriol Vinyals, and Quoc~V Le.
\newblock Sequence to sequence learning with neural networks.
\newblock \emph{Advances in neural information processing systems}, 27, 2014.

\bibitem[Vaswani et~al.(2017)Vaswani, Shazeer, Parmar, Uszkoreit, Jones, Gomez,
  Kaiser, and Polosukhin]{vaswani2017attention}
Ashish Vaswani, Noam Shazeer, Niki Parmar, Jakob Uszkoreit, Llion Jones,
  Aidan~N Gomez, {\L}ukasz Kaiser, and Illia Polosukhin.
\newblock Attention is all you need.
\newblock \emph{Advances in neural information processing systems}, 30, 2017.

\bibitem[Vincent et~al.(2008)Vincent, Larochelle, Bengio, and
  Manzagol]{vincent2008extracting}
Pascal Vincent, Hugo Larochelle, Yoshua Bengio, and Pierre-Antoine Manzagol.
\newblock Extracting and composing robust features with denoising autoencoders.
\newblock In \emph{Proceedings of the 25th international conference on Machine
  learning}, pages 1096--1103, 2008.

\bibitem[Wang and Cho(2019)]{wang2019bert}
Alex Wang and Kyunghyun Cho.
\newblock Bert has a mouth, and it must speak: Bert as a markov random field
  language model.
\newblock \emph{arXiv preprint arXiv:1902.04094}, 2019.

\bibitem[Welleck et~al.(2019)Welleck, Kulikov, Roller, Dinan, Cho, and
  Weston]{welleck2019neural}
Sean Welleck, Ilia Kulikov, Stephen Roller, Emily Dinan, Kyunghyun Cho, and
  Jason Weston.
\newblock Neural text generation with unlikelihood training.
\newblock \emph{arXiv preprint arXiv:1908.04319}, 2019.

\bibitem[Welleck et~al.(2020)Welleck, Kulikov, Kim, Pang, and
  Cho]{welleck2020consistency}
Sean Welleck, Ilia Kulikov, Jaedeok Kim, Richard~Yuanzhe Pang, and Kyunghyun
  Cho.
\newblock Consistency of a recurrent language model with respect to incomplete
  decoding.
\newblock \emph{arXiv preprint arXiv:2002.02492}, 2020.

\bibitem[Wolf et~al.(2020)Wolf, Debut, Sanh, Chaumond, Delangue, Moi, Cistac,
  Rault, Louf, Funtowicz, Davison, Shleifer, von Platen, Ma, Jernite, Plu, Xu,
  Scao, Gugger, Drame, Lhoest, and Rush]{wolf-etal-2020-transformers}
Thomas Wolf, Lysandre Debut, Victor Sanh, Julien Chaumond, Clement Delangue,
  Anthony Moi, Pierric Cistac, Tim Rault, Rémi Louf, Morgan Funtowicz, Joe
  Davison, Sam Shleifer, Patrick von Platen, Clara Ma, Yacine Jernite, Julien
  Plu, Canwen Xu, Teven~Le Scao, Sylvain Gugger, Mariama Drame, Quentin Lhoest,
  and Alexander~M. Rush.
\newblock Transformers: State-of-the-art natural language processing.
\newblock In \emph{Proceedings of the 2020 Conference on Empirical Methods in
  Natural Language Processing: System Demonstrations}, pages 38--45, Online,
  October 2020. Association for Computational Linguistics.
\newblock URL \url{https://www.aclweb.org/anthology/2020.emnlp-demos.6}.

\bibitem[Woolfson(2021)]{woolfson2021brief}
Derek~N Woolfson.
\newblock A brief history of de novo protein design: minimal, rational, and
  computational.
\newblock \emph{Journal of Molecular Biology}, 433\penalty0 (20):\penalty0
  167160, 2021.

\bibitem[Yang et~al.(2019)Yang, Wu, and Arnold]{yang2019machine}
Kevin~K Yang, Zachary Wu, and Frances~H Arnold.
\newblock Machine-learning-guided directed evolution for protein engineering.
\newblock \emph{Nature methods}, 16\penalty0 (8):\penalty0 687--694, 2019.

\bibitem[Yi et~al.(2021)Yi, Cho, and Bonneau]{yi2021nettime}
Ren Yi, Kyunghyun Cho, and Richard Bonneau.
\newblock Nettime: Improving multitask transcription factor binding site
  prediction with base-pair resolution.
\newblock \emph{bioRxiv}, 2021.

\end{thebibliography}

\end{document}